%% file: jinns_mloss.tex
\documentclass[twoside,11pt]{article}

\usepackage{blindtext}

\usepackage[preprint]{jmlr2e}
\usepackage[dvipsnames]{xcolor}
\usepackage{hyperref}

\DeclareFixedFont{\ttb}{T1}{txtt}{bx}{n}{10} %
\DeclareFixedFont{\ttm}{T1}{txtt}{m}{n}{10}  %

\usepackage{color}
\definecolor{deepblue}{rgb}{0,0,0.5}
\definecolor{deepred}{rgb}{0.6,0,0}
\definecolor{deepgreen}{rgb}{0,0.5,0}

 \usepackage{amsmath}
\usepackage{listings}
\usepackage{float}

\newcommand\pythonstyle{\lstset{
language=Python,
basicstyle=\ttm,
morekeywords={self},              %
keywordstyle=\ttb\color{deepblue},
emph={MyClass,__init__},          %
emphstyle=\ttb\color{deepred},    %
stringstyle=\color{deepgreen},
frame=tb,                         %
showstringspaces=false,
breaklines=false,
commentstyle=\color[HTML]{228B22},
}}

\newfloat{lstfloat}{htbp}{lop}
\floatname{lstfloat}{Listing}

\lstnewenvironment{python}[1][]
{
\pythonstyle
\lstset{#1}
}
{}

\newcommand\pythonexternal[2][]{{
\pythonstyle
\lstinputlisting[#1]{#2}}}

\newcommand\pythoninline[1]{{\pythonstyle\lstinline!#1!}}

\usepackage{makecell}
\usepackage{multirow} 
\usepackage{booktabs,siunitx}
\usepackage{lastpage}
\jmlrheading{23}{2024}{1-\pageref{LastPage}}{1/21; Revised 5/22}{9/22}{21-0000}{Hugo Gangloff and Nicolas Jouvin}

\ShortHeadings{jinns: Physics-Informed Neural Networks in JAX}{Gangloff and Jouvin}
\firstpageno{1}

\begin{document}

\title{\texttt{jinns}: a JAX Library for Physics-Informed Neural Networks}

\author{\name Hugo Gangloff \email hugo.gangloff@inrae.fr\\
\name Nicolas Jouvin \email nicolas.jouvin@inrae.fr \\
\addr MIA Paris-Saclay, INRAE, AgroParisTech, 
Université Paris-Saclay\\
}
\editor{My editor}

\newcommand{\CommentNico}[1]{\textcolor{Mahogany}{Nicolas: #1}}
\newcommand{\CommentHugo}[1]{\textcolor{OliveGreen}{Hugo: #1}}

\maketitle

\begin{abstract}%
\texttt{jinns} is an open-source Python library for physics-informed neural networks, built to tackle both forward and inverse problems, as well as meta-model learning. Rooted in the JAX ecosystem, it provides a versatile framework for efficiently prototyping real-problems, while easily allowing extensions to specific needs. Furthermore, the implementation leverages existing popular JAX libraries such as \texttt{equinox} and \texttt{optax} for model definition and optimisation, bringing a sense of familiarity to the user. Many models are available as baselines, and the documentation provides reference implementations of different use-cases along with step-by-step tutorials for extensions to specific needs. The code is available on Gitlab \url{https://gitlab.com/mia_jinns/jinns}.
\end{abstract}


\section{Introduction}
In the past decade, physics-informed neural networks \citep[PINNs,][]{raissi2019physics} have grown into an important area of machine learning, with many applications ranging from physics \cite{karniadakis2021physics}, to biology and ecology \cite{daneker2023systems, reyes2024spatio}. The general aim is to solve ordinary or partial differential equations with neural networks, by minimizing a loss accounting for the network's compliance to the prescribed dynamic. The differential operator governing the dynamic can be given and fixed or unknown, respectively leading to so-called \textit{forward} and \textit{backward} problems. One might also seek to solve an entire family of differential equations as in meta-modeling and operator learning \citep{li2020fourier}. The quick development of the literature has led to a myriad of research directions, from comparison to state-of-the-art numerical solvers to applications in real-world problems~\citep{cuomo2022scientific}. For the general mathematical context we refer to Appendix~\ref{app:context}.

 With this increased interest comes the need for efficient, open-source and reproducible software. Currently, several libraries have been deployed in different scientific programming languages and paradigms. We mention here, non-exhaustively, in both Julia and Python: the \texttt{NeuralPDE} \citep{neuralPDEjl}, \texttt{Modulus} \citep{modulus}, \texttt{PINA}~\citep{Coscia2023}, \texttt{IDRLnet}~\citep{peng2021idrlnet} and \texttt{Elvet}~\citep{araz2021elvet}. At the time of writing, one of the most popular Python library is \texttt{DeepXDE}~\citep{lu2021deepxde} that supports multiple backends: Tensorflow, PyTorch, Paddle and JAX.
 
Introduced a few years ago, the JAX Python library~\citep{jax2018github} stands out both for its computational performances, leveraging on \textit{just-in-time} (JIT) compilation, and its functional programming style. This allows for effortless vectorization and intuitive automatic differentiation in both forward and reverse modes, which is especially relevant when implementing partial derivatives. JAX has now grown into a large ecosystem for machine learning, with reference libraries for neural networks or optimization. 
 However, there is no dedicated JAX libraries for PINNs, and, due to JAX technical constraints, \texttt{DeepXDE} cannot be used to its full extent with this backend, especially for inverse problems. Aiming at filling this gap, \texttt{jinns}
 is a JAX-library focusing on inverse problems and meta-modeling. To the best of our knowledge, it is the first library for PINNs built solely on
 the JAX ecosystem, and providing hands-on implementation of refined PINNs architectures such as HyperPINNs or SeparablePINNs~\citep{de2021hyperpinn,cho2024separable}. %
\begin{figure}[!ht]
    \centering
    \includegraphics[width=.95\linewidth]{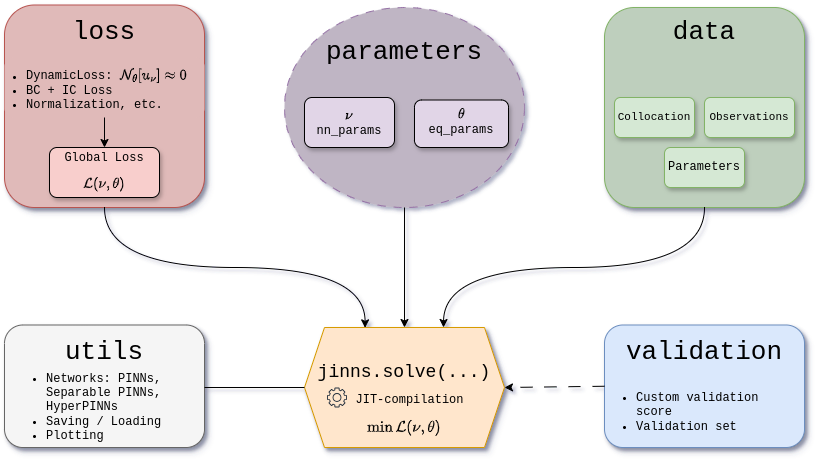}
    \caption{Typical user workflow for \texttt{jinns} users.}
    \label{fig:diagram-jinns}
\end{figure}

\section{Design and implementation}

\subsection{Organisation of the library}
\texttt{jinns} is a Python library with compatibility for Python 3.11+ built on-top of the JAX ecosystem. It is focused on user-flexibility and reproducibility. As such, it provides a modular interface allowing to specify the learning problem, defining a loss, a set of training points, and the parameters of interest to be optimized. Importantly, \texttt{jinns} relies on pure JAX, without any internal redefinition of data structures and transformations, so that users may directly use, \emph{e.g.}, \texttt{array} structures and \texttt{jit} or \texttt{grad} functions. The library is organized in modules dedicated to parts of the problem definition: the loss, the parameters, data generation, etc. Below is an overview of the different modules:

\begin{itemize}
    \item \textbf{data} -- This module handles the collocation points of the differential equation, as well as parameters and observed data for meta-modeling or inverse problems. Basic geometries such as cubic domains are implemented and users can define their own, or even use a pre-defined set of points, such as a pre-defined mesh. This module provides classes that can be thought of as equivalent to dataloaders.
    \item \textbf{parameters} -- This module provides the \texttt{Params} class, a specific module for the parameters of the problem. We explicitly distinguish between equation parameters, defining the differential operator, and the neural network's weights and biases, allowing to handle different scenarios of interest. In addition, an optional \texttt{DerivativeKeys} mask can be provided specifying how the different loss components should be differentiated with respect to the parameters.
    \item \textbf{loss} -- This module provides an API to define PDE-related loss (\texttt{DynamicLoss}) of the problem. The package handles vectorization over batch dimensions, hence users only have to implement the differential operator at one collocation points, as if they were writing the mathematical equation. In addition, this module implements standard differential operators such as divergence or Laplacian, as demonstrated in Appendix~\ref{app:loss_impl}. Both forward or reverse differentiation modes are available, allowing speed-ups based on the PINN architecture. Then, the global loss can be constructed, including initial or boundary conditions, and observations if any. A default class is provided for each type of differential equation, along with popular equations readily available in the package. 
    \item \textbf{utils} -- This module provides utility functions to create neural networks frequently used in the PINN litterature such as vanilla PINNs,\footnote{This corresponds to a multi-layer perceptron, where the input layer dimension depends on the differential equation $t$, $x$ or $(t, x)$.} Separable PINNs \citep[SPINN,][]{cho2024separable} or HyperPINNs \citep{de2021hyperpinn}. Importantly, \texttt{jinns} does not implement its own neural network module, but rather relies on the \texttt{equinox} package and its \texttt{Module} class. Hence, users may design their own architecture in a transparent fashion.
\end{itemize}

\subsection{Basic usage}

Figure~\ref{fig:diagram-jinns} describes \texttt{jinns}' workflow which breaks out as follows:
 
\paragraph{Defining the problem} -- First, users have to define their problem, specifying the data generation mechanism for collocation points, and possibly for observed data points. Then, users need to implement the differential equation, as well as the boundary and initial conditions, if any.

\paragraph{Defining the parameters} -- In \texttt{jinns}, parameters of interests are explicitly divided into two types: the neural network weights and biases $\nu$ and the differential operator parameters $\theta$. For advanced problems, the custom class \texttt{jinns.parameters.DerivativeKeys} let users specify a derivative mask for the parameters. The latter allows complete granularity as to which term of the loss should be differentiated with respect to which parameters. For example, in some inverse problems, one could need to differentiate the physics-term of the loss only with respect to the \texttt{nn\_params} ($\nu$), the \texttt{eq\_params} ($\theta$), or both. Finally,  equation parameters may depend on both time and space in a transparent fashion, which is handy for inhomegeneous equations.

\paragraph{Solving} -- The \texttt{jinns.solve} function performs the main optimization loop with respect to the parameters, handling forward and inverse problems as well as meta-modeling in a common API, and optionally using derivative masks specified by the user. Optimization in itself relies on \texttt{optax}, the reference JAX library for gradient-based optimization, and users are free to use any optimizer. The parameters and each term composing the loss may be monitored independently, facilitating diagnostic. Additionally, a custom validation step can be implemented separately and called periodically during training.

\subsection{Extensions}

The package is designed to be modular, and abstract classes are available for users to define their own data generation scheme, losses, PINN architectures, validation criterion, etc. On the technical side, JAX transforms such as \texttt{jit} or \texttt{grad} requires custom objects to be \textit{registered as PyTrees}. In \texttt{jinns}, this is done by inheriting from the \texttt{equinox.Module} class which handles PyTree registration in a transparent fashion, and any extension or custom implementation should be compatible with JAX transformations.

\subsection{Code quality} The source code follows best practices for Python software development and PEP 8 conventions. All custom classes and methods are documented and systematic type hinting improves readability of the code. 

\paragraph{Documentation} The package comes with a static documentation website for its API at \url{https://mia_jinns.gitlab.io/jinns/}, automatically updated by the Gitlab pipeline after each version update. In addition, tutorials are available for many aspects of the library, from introductory to advanced. Finally, we provide dozens of Jupyter notebook examples reproducing results from the literature.

\paragraph{Testing} The development takes place on Gitlab, with a read-only Github mirror for software dissemination. The source code is thoroughly tested by unit testing triggered by each merge request in our continuous integration pipeline on Gitlab. 

\subsection{Benchmark}
\label{sec:benchmark}
We used a community established benchmark \citep[PINNacle,][]{PINNacle} to compare \texttt{jinns} with the following popular frameworks: \texttt{DeepXDE}, \texttt{PINA} and \texttt{Modulus}. Timing and estimation performances can be found in App.~\ref{app:benchmark}, as well as technical details. Overall, jinns compares favorably to the other libraries, in particular, it is the fastest library for inverse problems. The reproducible benchmark code is available as a dedicated Gitlab repository: \url{https://gitlab.com/mia_jinns/pinn-multi-library-benchmark}.

\section{Conclusion}
 \texttt{jinns} is a fast growing JAX library for physics-informed machine learning problems, with a strong focus on providing a flexible and modular interface for inverse problems and meta-modeling. It compares favorably to existing libraries on community established benchmark problems .

\acks{
The authors would like to thank Lucia Clarotto, Pierre Gloaguen, Jean-Benoist Léger, Inass Soukarieh, and Achille Thin for their inputs on several aspects of the project.}

\appendix

\section{Context}
\label{app:context}
Physics-informed machine learning problems are described by a differential operator $\mathcal{N}_\theta$, defining the differential equation with parameters $\theta$. The goal is then to learn the weights and biases $\nu$ of a neural network so that it respects the prescribed equation. This can be summarized as a regression problem $\mathcal{N}_\theta[u_\nu] \approx 0$, where the residuals of the differential operator are computed on so-called collocation points. The latter can be sampled on the domain, with uniform or more refined scheme, or even prescribed by users beforehand. Additionally, one could have initial or boundary conditions, as well as noisy observations of the solution $u$ at some locations to be fitted. All this constraints are included in a global loss $\mathcal{L}(\nu, \theta)$ to be minimized. Then, there are several scenarios of interests: forward problems consist in minimizing the loss with respect to the network parameters $\nu$ for given equation parameters $\theta$, inverse problems seek to find the best set of parameters $\theta$. In addition, meta-modeling focuses on the task of learning a network $u_\nu$ outputting a solution for a family of parametric PDEs indexed by $\theta$.

\section{PDE-related loss implementation in \texttt{jinns}}
\label{app:loss_impl}
In this section, we illustrate the implementation of the PDE residuals in \texttt{jinns} on three different problems: a linear first order differential equation, a Poisson equation and a Fisher-KPP equation. The code uses JAX's functional programming style, with readily available automatic differentiation. Thanks to \texttt{jinns} internal handling of vectorization via \texttt{jax.vmap}, there is no need to handle the batch dimension so that users may focus on the PDE residuals at one point of the domain. Note that the class implementing the physics-related losses inherits from one of the three : \texttt{jinns.loss.ODE}, \texttt{jinns.loss.PDEStatio} or \texttt{jinns.loss.PDENonStatio} classes, which are internal abstract Python \texttt{dataclass} registered as PyTrees thanks to \texttt{equinox.Module}.

\subsection{Linear first order differential equation}
In Code~\ref{list:ode}, we show the definition in \texttt{jinns} of the dynamic loss in the case of a linear first order differential equation. The mathematical problem reads:
\begin{equation}
    \forall t \in [0, T], \quad \frac{\mathrm{d}}{\mathrm{d} t} u(t) - a u(t) = 0,
\end{equation}
with a coefficient $\theta = a  \in \mathbb R$.
\begin{lstfloat}
\pythonexternal[label={list:ode}, caption={Implementation of a linear first order differential equation in \texttt{jinns}.}, captionpos=b]{code/linear_ode.py}
\end{lstfloat}

\subsection{2D Poisson equation}
In Code~\ref{list:poisson}, we show the definition, in \texttt{jinns}, of the PDE-related loss in the case of the 2D Poisson equation. Recall that the mathematical problem reads:

\begin{equation}
   \forall (x,y) \in \mathbb{R}^2, \quad -\nabla \cdot (a(x,y)\nabla u(x,y)) = f(x, y),
\end{equation}
where $a$ is an heterogeneous diffusion coefficient varying in space.
\begin{lstfloat}
\pythonexternal[label={list:poisson}, caption=Implementation of a Poisson equation in \texttt{jinns}., captionpos=b]{code/poisson.py}
\end{lstfloat}
Note that thanks to \texttt{jinns} internal loss abstract definitions, the heterogeneous diffusion coefficient $a(x,y)$ is simply defined via the attribute \texttt{eq\_params\_heterogeneity} (see the \texttt{jinns} documentation for more details).

\subsection{Fisher KPP}
In Codes~\ref{list:kpp_reverse} and~\ref{list:kpp_fwd}, we show the definition, in \texttt{jinns}, of the PDE-related loss in the case of a Fisher KPP problem in arbitrary dimension (\texttt{dim\_x}), using both forward and reverse automatic differentiation. The forward case would for example match the SPINN models while the reverse case is the loss associated to vanilla PINN models.
Recall that the mathematical problem reads:
\begin{equation}
    \frac{\partial}{\partial t} u(t,x)=D\Delta u(t,x) + u(t,x)(r - \gamma u(t,x)),
\end{equation}
where $D$, $r$ and $\gamma$ are, respectively, the diffusion coefficient, the growth rate and the competition-effect parameters. Those parameters could easily be defined as functions depending on the space, time or other covariates (see previous section).
\begin{lstfloat}[p]
\pythonexternal[label={list:kpp_reverse}, caption={Implementation of a Fisher KPP PDE in \texttt{jinns} using reverse-mode automatic differentiation.}, captionpos=b]{code/fisherkpp_reverse.py}
\end{lstfloat}
\begin{lstfloat}[p]
\pythonexternal[label={list:kpp_fwd}, caption={Implementation of a Fisher KPP PDE in \texttt{jinns} using forward-mode automatic differentiation. Depending on the chosen architecture for \texttt{u}, \textit{e.g.} separable PINN, this can lead to important computational gains.}, captionpos=b]{code/fisherkpp_forward.py}
\end{lstfloat}

\section{PINN multi-library benchmark}
\label{app:benchmark}
\subsection{Description}
We propose a short benchmark study comparing \texttt{jinns} and three other popular PINN libraries in Python: \texttt{DeepXDE}~\citep{lu2021deepxde}, \texttt{PINA}~\citep{Coscia2023} and \texttt{Modulus}~\citep{modulus}. Several benchmark studies were recently proposed in the PINN litterature, such as PDEBench~\citep{PDEBench} and PINNacle~\citep{PINNacle}. While the latter addresses an extensive list of PDE problems using a variety of PINN models, the authors only used the \texttt{DeepXDE} library. To the best of our knowledge, this is the first study comparing several libraries.

We consider a selected list of PDE problems, addressing both forward and inverse problems, with an experimental set-up described in the following section. From PINNacle~\citep{PINNacle}, we reproduce 
\begin{itemize}
    \item \textbf{Burgers1D} -- Section 1 of Appendix B of the paper,
    \item \textbf{2D Navier-Stokes lid-driven flow} -- Section 11 of Appendix B of the paper,
    \item \textbf{Poisson Inverse} -- Section 21 of Appendix B of the paper.
\end{itemize}
From the \texttt{DeepXDE} documentation\footnote{\url{https://deepxde.readthedocs.io/en/latest/demos/pinn_inverse.html}} we reproduce
\begin{itemize}
    \item \textbf{Diffusion-Reaction inverse} -- Described at \url{https://deepxde.readthedocs.io/en/latest/demos/pinn_inverse/reaction.inverse.html},
    \item \textbf{Navier-Stokes inverse} -- Described in \citet[Section 4.1.1]{raissi2019physics}.
\end{itemize}
 All the code and instructions for reproducing our experiments described are available on a standalone repository at \url{https://gitlab.com/mia_jinns/pinn-multi-library-benchmark}. Contributions are appreciated and encouraged to enrich and improve this benchmark study.

\subsection{Experimental set-up}
Experiments were carried out on a local laptop equipped with a Nvidia T600 GPU. All hyper-parameters are equal across all libraries and clearly indicated at the beginning of each script. We only compared performances using a vanilla PINN architecture (multi-layered Perceptron), a fixed optimizer, and using regularly spaced collocation points over the domain, except for \texttt{Modulus}, where this option does not exist and collocation points are sampled from a uniform distribution. We emphasize that the original PINNacle benchmark compares different architectures, sampling schemes, optimizers and even loss function, which is beyond the scope of this short study here.

\subsection{Results}
Table~\ref{table:timing} presents training time (in seconds), while Tables~\ref{table:l1re} and~\ref{table:l2re} gives the L1 and L2 \textit{relative} error respectively. For forward problems, the error metrics are computed between the estimated PINN $\hat{u}$ and the reference solution $u^\star$ on a set of validation points. In inverse problems, we report the estimation error on the equation parameters $\hat{\theta}$ and $\theta^\star$. For the two last lines, the error are reported for the two scalar parameters of interested, hence the identical lines in Tables~\ref{table:l1re} and~\ref{table:l2re}.   

Looking at the results, some comments are in order. First of all, all experiments are reproducible by all libraries, with the notable exception of \texttt{DeepXDE} JAX backend which cannot handle inverse problems. Estimation errors vary between methods, which is a consequence of backend and implementation differences, but overall they all remain on the same scale, highlighting the consistency of each implementation. Second, \texttt{jinns} seems to be the fastest library for PINN modeling, especially when it comes to inverse problems with up to one order of magnitude ($\times 10$) speed-up for computations. Third, concerning implementation, each library has its own strength. For example, we found that \texttt{DeepXDE} has the most concise way of defining problems; while \texttt{jinns} allows more granularity on the optimization, especially with its \texttt{DerivativeKeys} option.

Finally, on a technical note, it is interesting to notice that we followed the experimental setup of PINNacle where mini-batching, either on collocation points, observations, or both, and stochastic optimization were not used. Such a feature is available in all libraries and could be useful to improve some timings and estimated solutions.
\input{tables/timing}

\input{tables/l1re}
\input{tables/l2re}

\newpage
\vskip 0.2in
\bibliography{jinns_mloss}

\end{document}

%% file: tables/timing.tex
\begin{table}[b]
\centering
\begin{tabular}{lccccc}
\toprule
& \multirow{2}{*}{\texttt{jinns}}& \multicolumn{2}{c}{DeepXDE} & \multirow{2}{*}{PINA}& \multirow{2}{*}{\makecell{NVIDIA\\ Modulus}}\\
\cmidrule(lr){3-4}
&  & JAX & Pytorch &  & \\ \midrule
Burgers1D    & \textbf{445}         &   723    &        671           &  1977    & 646 \\
NS2d-C            &   \textbf{265}    &     278     &   441            &  1600  & 275\\
PInv                   &   149    &          218     &        \texttt{CC}       & 1509 & \textbf{135} \\
Diffusion-Reaction-Inv &   \textbf{284}    & \texttt{NI}             &     3424   &  4061 & 2541\\
Navier-Stokes-Inv      &  \textbf{175}     & \texttt{NI}             &   1511         &   1403 & 498\\ \bottomrule
\end{tabular}
\caption{Training time in seconds on a Nvidia T600  GPU. \texttt{NI} means problem cannot be implemented in the backend, \texttt{CC} means the code crashed.}
\label{table:timing}
\end{table}

%% file: tables/l1re.tex
\begin{table}[p]
\centering
\begin{tabular}{lccccc}
\toprule
& \multirow{2}{*}{\texttt{jinns}}& \multicolumn{2}{c}{DeepXDE} & \multirow{2}{*}{PINA}& \multirow{2}{*}{\makecell{NVIDIA\\ Modulus}}\\
\cmidrule(lr){3-4}
&  & JAX & Pytorch &  & \\ \midrule
Burgers1D              &   0.013    &      \textbf{0.011}        &  0.012   &   0.040  &  0.019        \\
NS2d-C    &   \textbf{0.031}         &    0.064   &  0.070         &   0.385   & 0.076 \\ 
PInv                   &   0.070    &       0.170        &      \texttt{CC}          &  0.067 & \textbf{0.034}\\
Diffusion-Reaction-Inv &  $(0.016,\mathbf{0.013})$     & \texttt{NI}             &   $(\mathbf{0.005},0.017)$     &   $(0.024,0.028)$ &  $(0.018, 0.084)$   \\ 
Navier-Stokes-Inv   & $(0.007, 0.015)$    & \texttt{NI}  &  $(\mathbf{0.00},\mathbf{0.010})$ & $(0.001, 0.054)$& $(0.024, 0.027)$\\ \bottomrule
\end{tabular}
\caption{L1 relative error with respect to ground truth. \texttt{NI} means problem cannot be implemented in the backend, \texttt{CC} means the code crashed.}
\label{table:l1re}
\end{table}

%% file: tables/l2re.tex
\begin{table}[p]
\centering
\begin{tabular}{lccccc}
\toprule
& \multirow{2}{*}{\texttt{jinns}}& \multicolumn{2}{c}{DeepXDE} & \multirow{2}{*}{PINA}& \multirow{2}{*}{\makecell{NVIDIA\\ Modulus}}\\
\cmidrule(lr){3-4}
&  & JAX & Pytorch &  & \\ \midrule
Burgers1D              &  0.050     &       0.035        &   \textbf{0.020}    &       0.147 &   0.101  \\
NS2d-C   & \textbf{0.064}       &    0.105           &     0.105        &    0.623 & 0.125\\
PInv                   &  0.088     &       0.262        &     \texttt{CC}     &    0.083  & \textbf{0.043} \\
Diffusion-Reaction-Inv &  $(0.016,\mathbf{0.013})$     & \texttt{NI}             &   $(\mathbf{0.005},0.017)$     &   $(0.024,0.028)$ &  $(0.018, 0.084)$   \\ 
Navier-Stokes-Inv   & $(0.007, 0.015)$    & \texttt{NI}  &  $(\mathbf{0.00},\mathbf{0.010})$ & $(0.001, 0.054)$& $(0.024, 0.027)$\\ \bottomrule
\end{tabular}
\caption{L2 relative error with respect to ground truth.  \texttt{NI} means problem cannot be implemented in the backend, \texttt{CC} means the code crashed.}
\label{table:l2re}
\end{table}